\documentclass[runningheads]{llncs}
\usepackage{graphicx}
\usepackage{subfigure}
\usepackage{booktabs} % for professional tables
\usepackage{hyperref}
\usepackage{tabularx}
\usepackage{arydshln} % for dashed hlines (\hdashline)
\usepackage{amssymb}
\usepackage{amsmath}
\makeatletter
\let\oldtheequation\theequation
\renewcommand\tagform@[1]{\maketag@@@{\ignorespaces#1\unskip\@@italiccorr}}
\renewcommand\theequation{(\oldtheequation)}
\makeatother

\usepackage{nicefrac}       % compact symbols for 1/2, etc.
\usepackage{microtype}      % microtypography
\usepackage{nccmath}
\usepackage[dvipsnames]{xcolor}
\usepackage[binary-units=true]{siunitx}
\usepackage{multicol}
\usepackage[algo2e, ruled, lined, longend]{algorithm2e}
\usepackage{algorithm}
\usepackage[bottom]{footmisc}  % Footnotes should be below figures and tables
\usepackage[switch]{lineno}

\newcolumntype{P}[1]{>{\centering\arraybackslash}p{#1}}
\newcolumntype{R}{>{\raggedleft\arraybackslash}X}
\newcolumntype{C}{>{\centering\arraybackslash}X}

\SetCommentSty{mycommentfont}

\begin{document}
\title{Latent State Inference in a Spatiotemporal Generative Model\thanks{Funded by the Deutsche Forschungsgemeinschaft (DFG, German Research Foundation) under Germany’s Excellence Strategy - EXC number 2064/1 – Project number 390727645. Moreover, we thank the International Max Planck Research School for Intelligent Systems (IMPRS-IS) for supporting Matthias Karlbauer.}}
%
%\titlerunning{Abbreviated paper title}
% If the paper title is too long for the running head, you can set
% an abbreviated paper title here
%
\author{Matthias Karlbauer\inst{1}\orcidID{0000-0002-4509-7921} \and
Tobias Menge\inst{1} \and\\
Sebastian Otte\inst{1}\orcidID{0000-0002-0305-0463} \and
Hendrik P. A. Lensch\inst{2}\orcidID{0000-0003-3616-8668} \and
Thomas Scholten\inst{3}\orcidID{0000-0002-4875-2602} \and
Volker Wulfmeyer\inst{4}\orcidID{0000-0003-4882-2524} \and
Martin V. Butz\inst{1}\orcidID{0000-0002-8120-8537}
}
\authorrunning{Karlbauer et al.}
% First names are abbreviated in the running head.
% If there are more than two authors, 'et al.' is used.
%
\institute{
	University of T\"ubingen -- Neuro-Cognitive Modeling Group,\\
	Sand 14, 72076 T\"ubingen, Germany, \email{martin.butz@uni-tuebingen.de}
	\vspace{0.2cm} 
	\and
	University of T\"ubingen -- Computer Graphics,\\
	Maria-von-Linden-Stra\ss e 6, 72076 T\"ubingen, Germany \vspace{0.2cm} 
	\and
	University of T\"ubingen -- Soil Science and Geomorphology,\\
	R\"umelinstra\ss e 19-23, 72070 T\"ubingen, Germany \vspace{0.2cm} 
	\and
	University of Hohenheim -- Institute for Physics and Meteorology,\\
	Garbenstra\ss e 30, 70599 Stuttgart, Germany
	\vspace{-0.2cm}
}
\maketitle              % typeset the header of the contribution
\begin{abstract}
Knowledge about the hidden factors that determine particular system dynamics is crucial for both explaining them and pursuing goal-directed interventions.
Inferring these factors from time series data without supervision remains an open challenge.
Here, we focus on spatiotemporal processes, including wave propagation and weather dynamics, for which we assume that universal causes (e.g. physics) apply throughout space and time. 
A recently introduced DIstributed SpatioTemporal graph Artificial Neural network Architecture (DISTANA) is used and enhanced to learn such processes, requiring fewer parameters and achieving significantly more accurate predictions compared to temporal convolutional neural networks and other related approaches.
We show that DISTANA, when combined with a retrospective latent state inference principle called active tuning, can reliably derive location-respective hidden causal factors.
In a current weather prediction benchmark, DISTANA infers our planet's land-sea mask solely by observing temperature dynamics and, meanwhile, uses the self inferred information to improve its own future temperature predictions.

\end{abstract}
%
%
%
%%%%%%%%%%%%%%%%%%%%%%%%%%%%%%%%%%%%%%%%
%%%%%%%%%%%%% INTRODUCTION %%%%%%%%%%%%%
%%%%%%%%%%%%%%%%%%%%%%%%%%%%%%%%%%%%%%%%

\section{Introduction}

When considering our planet's weather, centuries of past research have identified a large number of factors that affect its highly nonlinear and partially chaotic dynamics.
Yet, can we ever be sure of having identified all hidden causal factors?
Moreover, do we have (sufficient) data about them?  
These are fundamental questions in any prediction or forecasting task, including spatiotemporal processes such as soil property dynamics, traffic forecasting, energy-flow prediction (e.g in brains or supply networks), or recommender systems.
Here, we investigate how unobservable hidden factors may be inferred from spatiotemporal data streams. 

When regularities in hidden causes are detectable, they may be encoded in the latent activities of recurrent neural networks \cite{Rabinowitz:2018,rodriguez2019modeling}, such as a long short-term memory (LSTM) \cite{Hochreiter:1997}.
The involved and conventional forward-directed inference of recurrent neural networks, however, has two main disadvantages:
First, the encodings of the hidden causes form while streaming data, meaning they are not available from the beginning of a sequence.
Second, learning, detecting and shaping the encodings is relatively hard, because the error signal only decreases once the unfolding data stream is suitably compressed.

To overcome these limitations, we combine and extend the recently introduced DIstributed SpatioTemporal graph Artificial Neural network Architecture (DISTANA) \cite{karlbauer2019distributed} with active tuning (AT) \cite{Butz:2019,Butz:2019a,otte2020active}, which facilitates the determination of hidden causal states via retrospective inference over time.
Projected onto stable neural states, akin to parametric bias neurons \cite{Tani:2004,Sugita:2011a}, AT searches for constant input biases, assuming that the observed dynamics are influenced by particular constant and only indirectly observable factors.

Following the idea of \emph{relational inductive biases} \cite{battaglia2018relational}, DISTANA is designed to model the hidden causal processes that generate spatiotemporal dynamics. 
Hence, DISTANA assumes that the sensed dynamics are generated by universal causal principles (e.g.\ physics).
Moreover, we endow DISTANA with the expectation that constant, hidden factors modify the spatiotemporal processes locally.
For example, weather dynamics follow the universal principles of thermodynamics from physics and are locally dependent on the topology.

The contributions we make are as follows: (A) the combination of DISTANA with active tuning (AT) to infer constant, hidden factors locally, even when these factors are never made available to the network  -- neither as input nor as (target) output. (B) we show that reasonable latent neural activities are inferred during training and testing via retrospective spatiotemporal analysis.
% (C) To the best of our knowledge, we are the first to apply graph neural networks to weather data \cite{wilson2018low} (at global scale), by using DISTANA to model global weather benchmark data \cite{rasp2020weatherbench}.
(C) after having learned a distributed, generative model of the globally unfolding dynamics, we demonstrate that our planet's land-sea mask as well as other causal factors can be inferred via the retrospective analysis of unfolding weather dynamics -- partially again even when the algorithm was never informed about these factors -- to increase the model's prediction abilities.

We conclude that the retrospective inference of latent states via AT offers a promising method to identify hidden factors in data streams, and that graph neural networks (GNN) like DISTANA bear great potential at modeling real-world spatiotemporal processes.
%In fact, we believe that the introduced principles are applicable to many other domains. 
%While our inductive learning bias essentially assumes the presence of locally and temporally constant causal factors, modifications of these assumptions are possible and easy to implement.

%%%%%%%%%%%%%%%%%%%%%%%%%%%%%%%%%%%%%%%%
%%%%%%%%%%%%%%% METHODS %%%%%%%%%%%%%%%%
%%%%%%%%%%%%%%%%%%%%%%%%%%%%%%%%%%%%%%%%

\section{DISTANA}
% We first introduce the considered benchmarks. Next, we detail the developed algorithms, focusing on DISTANA, its actual implementation in the light of the benchmarks, and the integration of AT.

\begin{figure}[t]
	\centering
	\includegraphics[width=\textwidth]{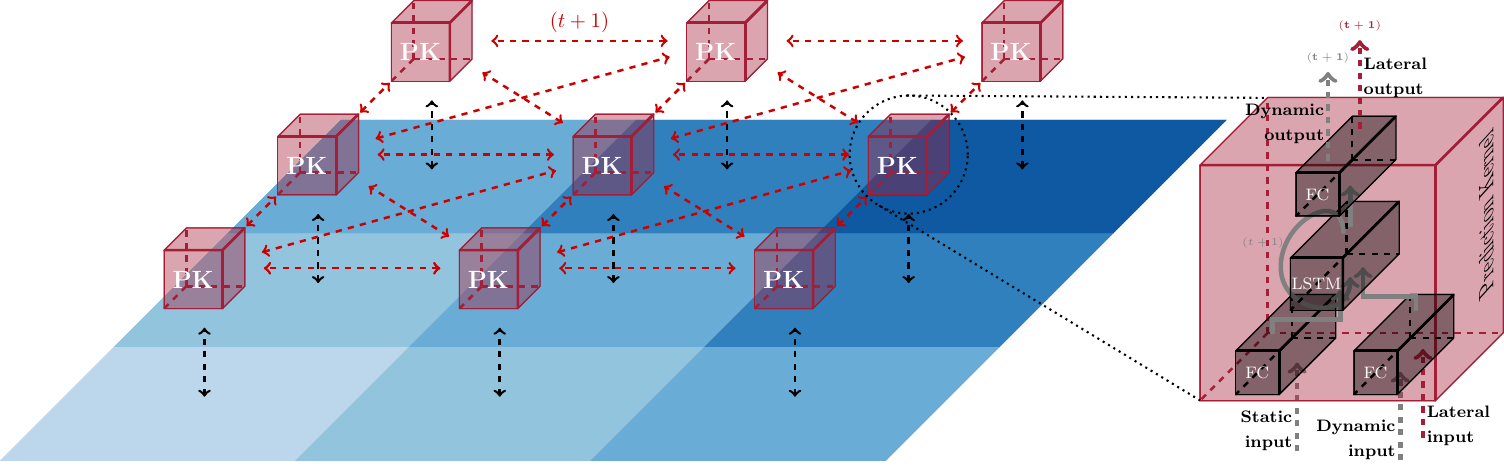}\hfill
	\caption{$3\times3$ sensor mesh grid showing the connection scheme of Prediction Kernels (PKs) that model the local dynamical process while communicating laterally. Figure modified from \cite{karlbauer2020inferring}.}
	\label{methods:fig:distana_architecture}
\end{figure}

As introduced in \cite{karlbauer2019distributed} and following the naming convention of \cite{wu2019comprehensive}, DISTANA (the DIstributed SpatioTemporal graph Artificial Neural network Architecture) can be described as a spatiotemporal graph neural network (ST-GNN).
While GNNs give the designer a large amount of freedom in controlling the flow of information within the model (referred to as relational inductive biases) \cite{battaglia2018relational}, they are reported to model physical systems with very high precision and accuracy for up to several hundreds of time steps even during closed loop prediction \cite{battaglia2016interaction,santoro2017simple,kipf2018neural,sanchez2018graph,van2018relational}.
%However, one hurdle of GNNs is the way the graph is structured, i.e., which vertex pairs are connected by edges to allow for information exchange.
Thorough surveys about GNNs and the numerous ways of creating the graph and setting up the connection schemes are written by \cite{bronstein2017geometric,battaglia2018relational,wu2019comprehensive}.

The GNN used in this work, DISTANA, consists of prediction kernels (PKs), which are arranged in a lattice structure.
PKs model local dynamics concurrently. 
In every time step $t$, each PK receives (i) local dynamic data, and (ii) lateral output activities from the neighboring PKs from $t-1$ to exchange information between PKs.
Here, we extend the PKs to (iii) additionally receive location specific static inputs.
The time recurrent PKs process this information, combine it with their previous latent state, and generate (i) predictions of the next local dynamic data input at $t+1$, as well as (ii) outputs to the laterally connected PKs (cf.  \autoref{methods:fig:distana_architecture}).
PKs are akin to a spatiotemporal convolutional kernel, since all PKs share identical weights, that is, a single set of weights is applied and optimized in every grid cell.
As a result, the likelihood of overfitting local data irregularities is reduced and the emergence of a highly generalizing and universally applicable set of weights is fostered.
Because of the reduction of trainable weights, less data is needed for training.

\subsection{Alternative State-of-the-Art Architectures}
\begin{figure*}[t]
	\centering
	\includegraphics[height=1.7cm]{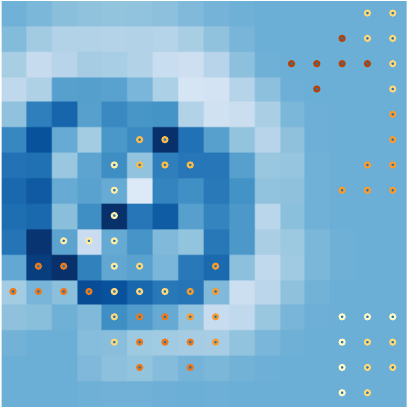}\hfill\includegraphics[height=1.7cm]{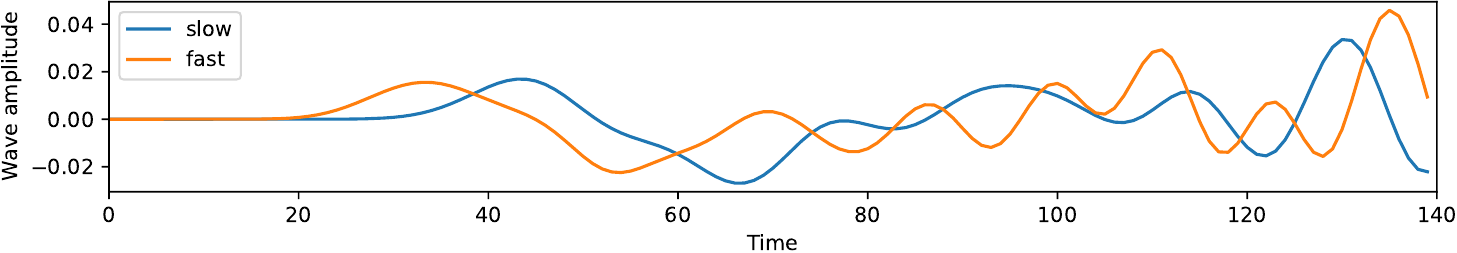}
	\caption{Left: two-dimensional wave propagating through a $16\times16$ grid with obstacles.
	Darker dots in the grid nodes correspond to stronger blocking effect on the wave.
	Right: wave activity for two exemplary positions in the grid, with fast and slow propagation speeds.}
	\label{methods:fig:circular_wave}
\end{figure*}

We compared DISTANA with two well-suited deep learning approaches.
First, we tested convolutional long short-term memory models (ConvLSTMs) \cite{xingjian2015convolutional} to predict circular wave dynamics (see \autoref{methods:fig:circular_wave}). The used ConvLSTM model has $2\,952$ free parameters to project the $16\times16\times1$ input (ignoring batch and time dimensions) via the first layer on eight feature maps (resulting in dimensionality $16\times16\times8$) and subsequently via the second layer back to one output feature map. 
All kernels have a filter size of $k = 3$, apply zero-padding and are implemented with a stride of one.
The code was taken and adapted from\footnote{\scriptsize\url{https://github.com/ndrplz/ConvLSTM_pytorch}}.
Second, we tested Temporal Convolution Networks (TCN) \cite{kalchbrenner2016neural,dauphin2017language,bai2018empirical}.
The TCN used in this work is a three-layer network with $2\,306$ parameters, where the input layer projects to eight feature maps, which project their values back to one output value.
A kernel filter size of $k = 3$ is used for the two spatial dimensions in combination with the standard dilation rate of $d = 1, 2, 4$ for the temporal dimension, resulting in a temporal horizon of 28 time steps, cf. \cite{bai2018empirical}.
Various experiments with other sizes and deeper network structures have not yielded any better performance than the one reported.
Code was taken and adapted from \cite{bai2018empirical}.

\subsection{Static Input Inference via Active Tuning}\label{methods:subsec:gradient_inference_technique}

Essentially, active tuning (AT) \cite{Butz:2019,Butz:2019a,otte2020active} can be seen as a different paradigm for handling RNNs: instead of the usual ${\text{input} \rightarrow \text{compute} \rightarrow \text{output}}$ scheme, a subset of the RNN's neurons is decoupled from the direct input signal.
The activation of this subset of neurons is computed from the RNN's prediction-based gradients, both during training and testing.
Gradient information is obtained from backpropagating the discrepancy between the RNN's predicted output $\mathbf{\hat{y}}$ and the desired output $\mathbf{y}$.
Thus, the neuron dynamics of the subset is solely influenced by the target indirectly, by means of temporal gradient information induced by the prediction error.

In this work, as mentioned before, DISTANA receives dynamic and lateral input, while the static input $\mathbf{s}$ is withheld and must be inferred via AT to reasonably model the unfolding dynamics. Technically, $\mathbf{s}$ is fed to the model initially as a zero vector and optimized iteratively through the AT method.
AT is applied to reduce local prediction errors, while the PK weights are updated as usual to reduce global prediction errors and model universal dynamics.

The active tuning algorithm, please refer to \cite{otte2020active} for more information, can be applied in combination with any desired gradient optimization strategy, e.g. Adam \cite{kingma2014adam}.
Furthermore, an arbitrary number of optimization cycles $c$, here $c=1$, and history length $H$, here $H=10$, can be chosen, where the latter indicates up to what time in the local past the latent context vector $\mathbf{s}$, which is assumed to be constant, is optimized.
The AT optimization procedure is realized every ten time steps retrospectively on the predicted dynamic input, starting from time step $\tau$ to find a converging $\mathbf{s}$.

%\begin{algorithm}[b!]
%	\SetNoFillComment
%	\DontPrintSemicolon
%	\SetKwInOut{KwIn}{Input}
%	\KwIn{$c, H, \tau, \mathbf{d}, \mathbf{\hat{d}}, \mathbf{l}, \mathbf{s}$}
	%\STATE{ActiveTuning}{($c, H, \tau, \mathbf{d}, \mathbf{l}, \mathbf{s}$)}
%	\BlankLine
%	\For(){$c$ in range(0, $c$)}{
%		\tcc{Compute and apply gradients on $\mathbf{s}$}
%		$\mathbf{s}\gets\mathbf{s} - \eta\cdot\text{gradients}(\text{mse}(\mathbf{d}, \mathbf{\hat{d}}))$
%		\vskip 2pt
%		\BlankLine
%		\For(){$t^\prime$ in range($\tau-H+1$, $\tau+1$)}{
%			\vskip 2pt
%			\tcc{Generate model prediction $\mathbf{\hat{d}}$}
%			$\mathbf{\hat{d}}^{t^\prime}\gets\text{model}(\mathbf{\hat{d}}^{t^\prime-1}, \mathbf{l}^{t^\prime-1}, \mathbf{s})$
%		}
%	}
%	\BlankLine
	%\STATE{Return }{$\mathbf{s}$}
%	\KwRet{$\mathbf{s}$}
%	\caption{Active Tuning}\label{methods:alg:active_tuning}
%\end{algorithm}

%\subsubsection{Modifications to Active Tuning}

We have modified (AT) for application on two-dimensional data in order to infer an individual, slowly changing local latent variable, denoted as $\mathbf{s}_i$, for each vertex of the two-dimensional grid.
AT so far has been applied to one-dimensional time series prediction for the inference of rarely changing contextual \cite{Butz:2019,Butz:2019a} or dynamically changing latent states \cite{otte2020active}.
In contrast to previous applications of AT, the inferred local static input $\mathbf{s}$ is not reset between sequences during training here, assuming a constant static context.% that is to be inferred during training and testing.

%Furthermore, in order to reduce high-frequency noise from the gradient signal being encoded in the inferred static input $\mathbf{s}$, a temporal low-pass filter was applied: ${\mathbf{\hat{s}}^t\leftarrow \alpha\cdot\mathbf{\hat{s}}^t + (1 - \alpha)\cdot\mathbf{\hat{s}}^{t-1}}$ with $\alpha=0.1$.
%Smaller values of $\alpha$ lead to both an increased resistance against noise and an increased system inertia, preventing it from adapting to only temporary signals.
%In our experiments, $alpha=0.1$ proved to be a suitable choice.
%For the inference of the static input with AT, a learning rate of $\eta=1e-3$ was chosen in combination with the number of optimization cycles $c=3$.
%While large values of $c$ increase the rate of convergence and susceptibility of the static input inference in each AT optimization step, the computational expenses increase too.

In our initial experiments, the inferred static input frequently drifted or potentially exploded, comparable to an Intern Covariate Drift \cite{pmlr-v37-ioffe15}.
We solve this problem, similarly to \cite{pmlr-v37-ioffe15}, by normalizing the inferred latent variable (in our case the static input $\textbf{s}$), via the mean $\mu_s$ and standard deviation $\sigma_s$ with respect to all inferred static inputs $\{s_i^t\}_{i=1}^k$, where $k$ is the number of cells or pixels in the two-dimensional field.
Additionally, to remove noise from the inference process caused by inconsistent gradient signals before and after the normalization, the weights $W$ and the bias neuron $b$ of the static input preprocessing layer are modified such that the activation of the static input preprocessing layer remains the same before and after the normalization:
\begin{equation}
b \leftarrow b + W \cdot \mu_s; \hspace{0.8cm} W \leftarrow W \cdot \sigma_s
\label{eq:AT-Redistribution}
\end{equation}
While the activation of the network is preserved, the gradients backpropagated through the static input preprocessing layer are affected asymmetrically by the modified weights and bias. In our experiments, this has been shown to substantially improve both the inference during training and the convergence of the inference process during testing.

%To smooth the development of the static context $\mathbf{s}$ via AT during training, in each iteration $t$, a temporal low pass filter was applied: ${\hat{\mathbf{s}}^t \leftarrow \alpha\cdot\hat{\mathbf{s}}^{t} + (1 - \alpha)\cdot\hat{\mathbf{s}}^{t-1}}$ with ${\alpha=0.05}$.
%Next, the static values were normalized (with respect to the mean $\mu_{\hat{\mathbf{s}}}$ and standard deviation $\sigma_{\hat{\mathbf{s}}}$ over all spatial locations):
%${\hat{\mathbf{s}} = (\hat{\mathbf{s}} - \mu_{\hat{\mathbf{s}}}) / \sigma_{\hat{\mathbf{s}}}}$.
%Finally, values were clipped according to a $2.5\sigma$-rule, with respect to moving average values of $\mu_{\hat{\mathbf{s}}}$ and $\sigma_{\hat{\mathbf{s}}}$, to suppress negative influences from outliers.
%Only with these crucial AT-modifications, a reasonable inference of static inputs could be achieved.

%%%%%%%%%%%%%%%%%%%%%%%%%%%%%%%%%%%%%%%%
%%%%%%%%%%%%% EXPERIMENTS %%%%%%%%%%%%%%
%%%%%%%%%%%%%%%%%%%%%%%%%%%%%%%%%%%%%%%%

\section{Experiments and Results}

The experiments are based on two classes of spatiotemporal time series.
Both are representatives of universal, but locally and temporally modifiable, spatiotemporal, causal processes that propagate dynamics over local topologies throughout a homogeneously connected graph.

\subsection{2D Circular Wave}\label{experiments:subsec:2d_circular_wave}

Following \cite{karlbauer2019distributed}, a spatiotemporal wave propagation dataset was created to validate our approach.
%More precisely, the two-dimensional wave equation was solved by means of the second order central differences method, resulting in an equation to calculate the elevation score for each position of the considered $16\times16$-pixel field in the next time step (see appendix in the supplementary material for details on the equations).
In comparison to \cite{karlbauer2019distributed}, however, the data generation was enhanced such that the wave propagation velocity could be contextually modified locally, which intuitively resembles obstacles in the water, which affect the wave's propagation behavior (cf. \autoref{methods:fig:circular_wave}).

This benchmark was used to (a) demonstrate and compare DISTANA's principal capability to model locally parameterized spatiotemporal dynamics and (b) determine whether DISTANA can be used in combination with AT to infer an underlying and hidden static (causal) factor, which modifies the observed dynamics locally.
Adam \cite{kingma2014adam} is used for training with a learning rate of $\SI{e-3}{}$ along with Scheduled Sampling \cite{bengio2015scheduled} with a linear slope of 270 epochs, transitioning from a probability of ${0.0\rightarrow0.9}$ of feeding the network with its own output in the next iteration instead of the teacher signal. During each sequence, 30 teacher forcing steps are conducted to induce reasonable network activities before switching to closed loop.
Network inputs $\mathbf{x}$ and the according targets $\mathbf{y}$ are exactly the same sequences shifted by one time step to train four different model types (ConvLSTM, TCN, DISTANA and DISTANA + AT) to iteratively predict the next two-dimensional dynamic wave field state (one step ahead prediction).
For the static input inference, DISTANA is augmented with a parametric bias neuron, whose activity is inferred during training and testing, aiming at the identification of an unknown location-specific wave velocity-influencing factor (static context).
Training was realized over $300$ epochs consisting of $100$ training sequences of length $120$ each.
The target static context vector $\mathbf{s}\in\mathbb{R}^{16\times16}$ was initialized by drawing values from $\{0.2, 0.3, 0.5, 0.6, 0.8, 0.9\}$, where small values cause the waves to propagate slower at the according pixel.
An exemplary ground truth context map $\mathbf{s}_{GT}$ is visualized in \autoref{methods:fig:circular_wave} (left, brownish dots).
Note that $\mathbf{s}_{GT}$ was used for the data generation but has never been provided to any model.
The preprocessing layer size of DISTANA was set to eight neurons and the subsequent LSTM layer consisted of twelve cells, yielding $1\,236$ parameters.
For the DISTANA + AT model, an additional static preprocessing layer with five neurons was used, resulting in $1\,486$ weights overall, compared to $2\,952$ and $2\,306$ weights for ConvLSTM and TCN, respectively.

To test the models' generalization capabilities, 16 new static context vectors $\mathbf{s}^\prime$ have been generated by drawing from $\{0.2, 0.3, \dots, 1.0\}$ (e.g. see \autoref{experiments_and_results:fig:context_inference}, top right-most).
All models were evaluated on 50 sequences -- made up of 120 time steps each -- per $\mathbf{s}^\prime$.
Reasonable activity was induced into the models by applying 30 steps of teacher forcing, followed by 90 steps of closed loop prediction for which an average MSE over all test examples and spatial locations was computed. 
For DISTANA + AT, the static context has been inferred before the testing on 50 separate sequences, using a history length of $H=30$, one optimization cycle ($c = 1$), and an inference learning rate of $\eta=0.1$ for the first three epochs, and $\eta=0.01$ for the remaining seven epochs.

\subsection{2D Circular Wave Results}

The prediction accuracy of ConvLSTM, TCN, DISTANA and DISTANA + AT differs considerably.
TCN without scheduled sampling tends to start oscillating increasingly after few steps of closed loop prediction, resulting in a mediocre MSE score of \SI{2.94(238)e-2}{}, while ConvLSTM \SI{8.69(87)e-4}{} and DISTANA \SI{8.69(87)e-4}{}, both trained with scheduled sampling, tend to vanish after few steps of closed loop prediction. Solely DISTANA + AT trained with scheduled sampling is able to preserver a stable activation pattern with an MSE of \SI{3.87(248)e-4}{}. %(\autoref{experiments_and_results:fig:context_inference}).

\begin{figure}[t]
	\centering
	\begin{minipage}{0.48\textwidth}
		\includegraphics[width=0.19\columnwidth]{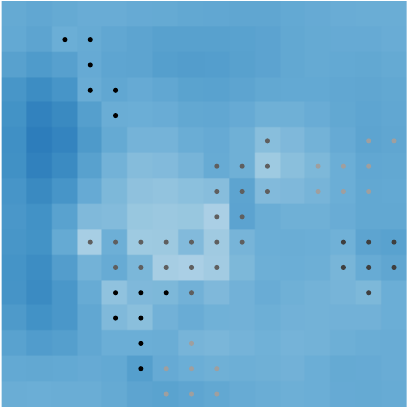}\hfill\includegraphics[width=0.19\columnwidth]{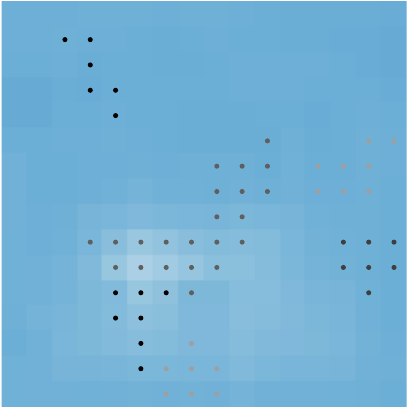}\hfill\includegraphics[width=0.19\columnwidth]{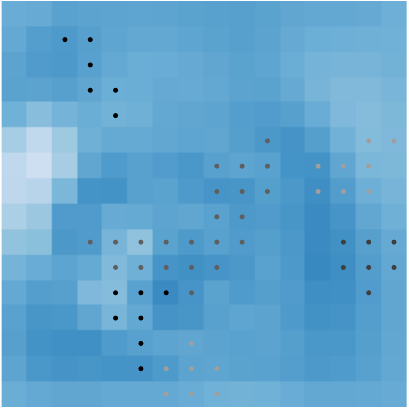}\hfill\includegraphics[width=0.19\columnwidth]{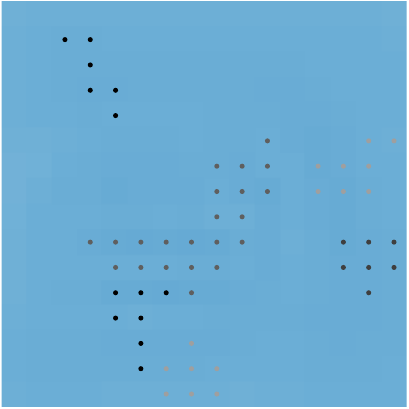}\hfill\includegraphics[width=0.19\columnwidth]{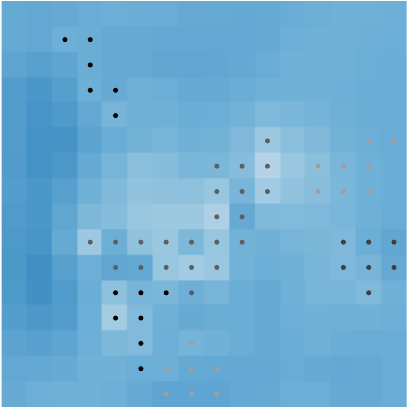}\\
		\vspace{-0.05cm}
		\includegraphics[width=0.99\columnwidth]{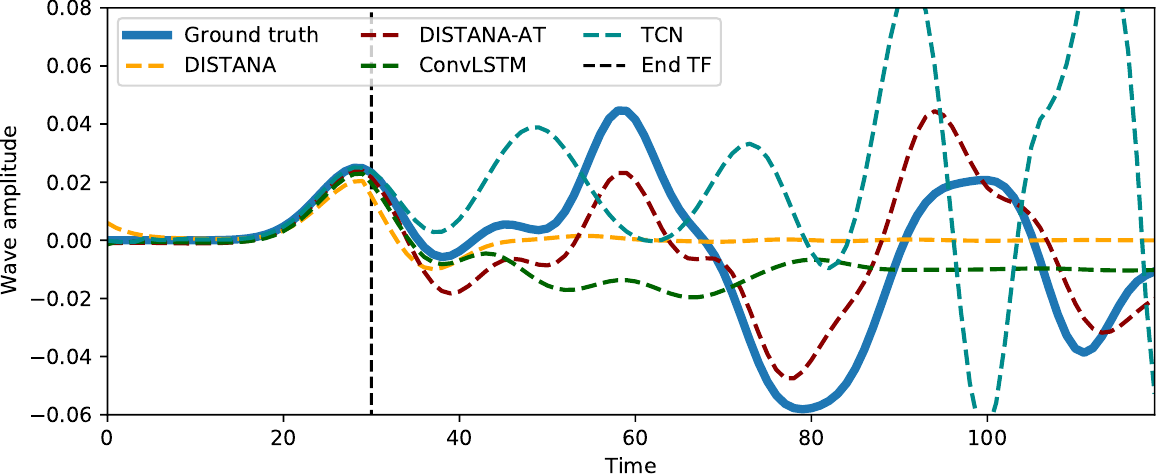}
	\end{minipage}
	\hfill
	\begin{minipage}{0.48\textwidth}
		\includegraphics[width=0.19\columnwidth]{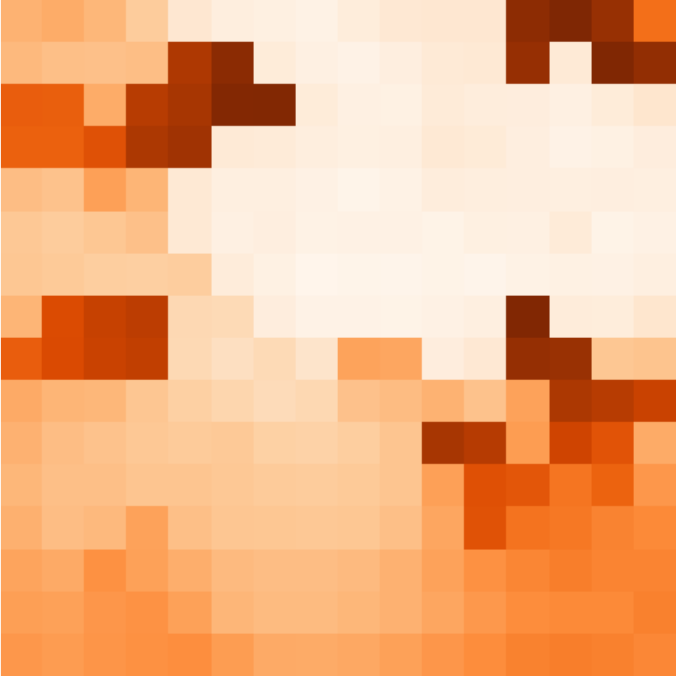}\hfill\includegraphics[width=0.19\columnwidth]{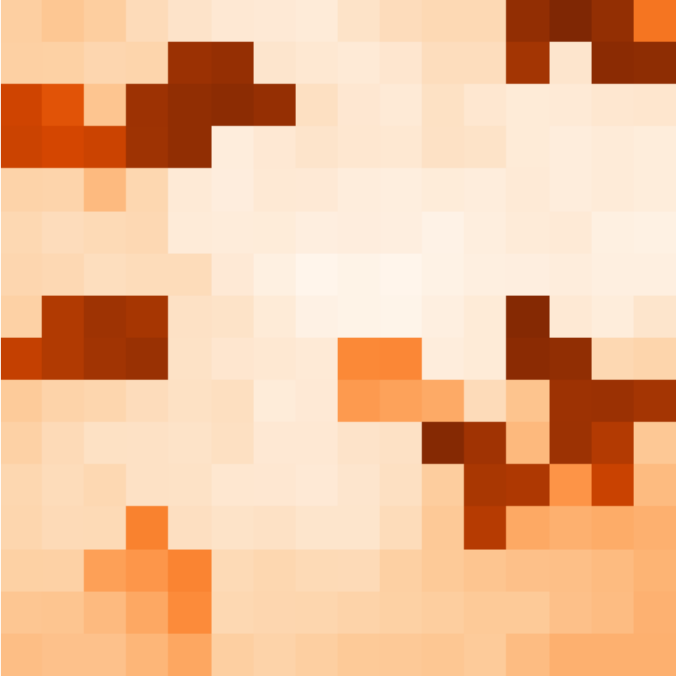}\hfill\includegraphics[width=0.19\columnwidth]{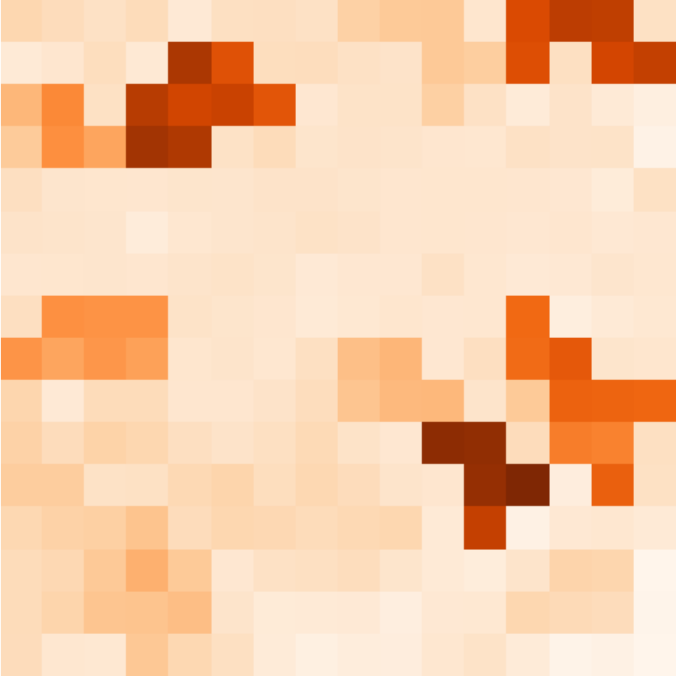}\hfill\includegraphics[width=0.19\columnwidth]{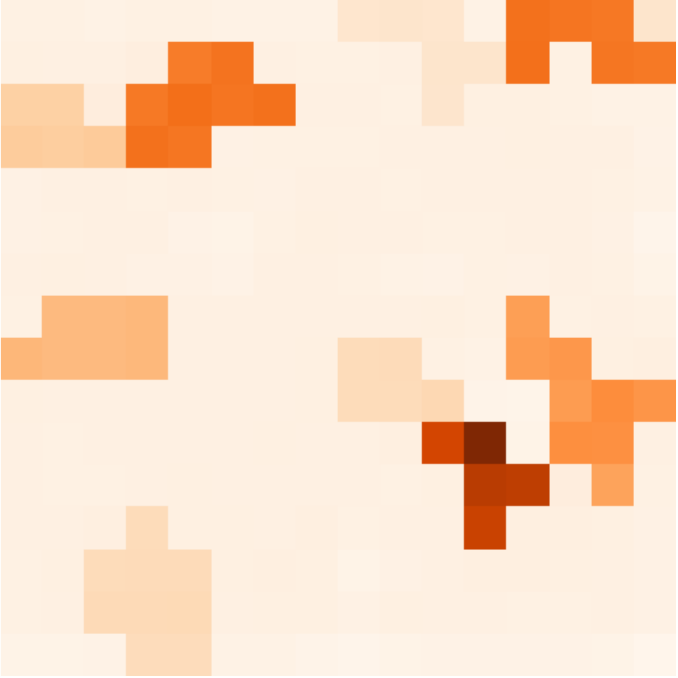}\hfill\includegraphics[width=0.19\columnwidth]{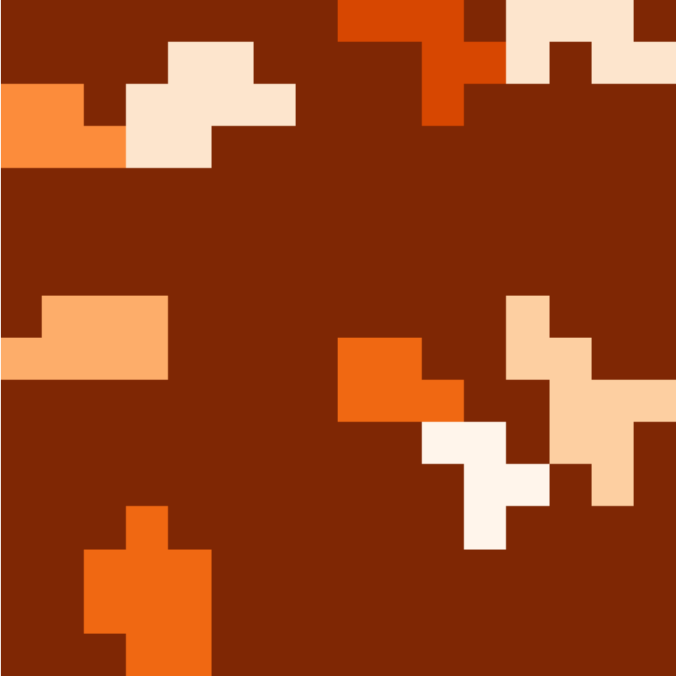}\\
		\vspace{-0.05cm}
		\hfill\includegraphics[width=0.99\columnwidth]{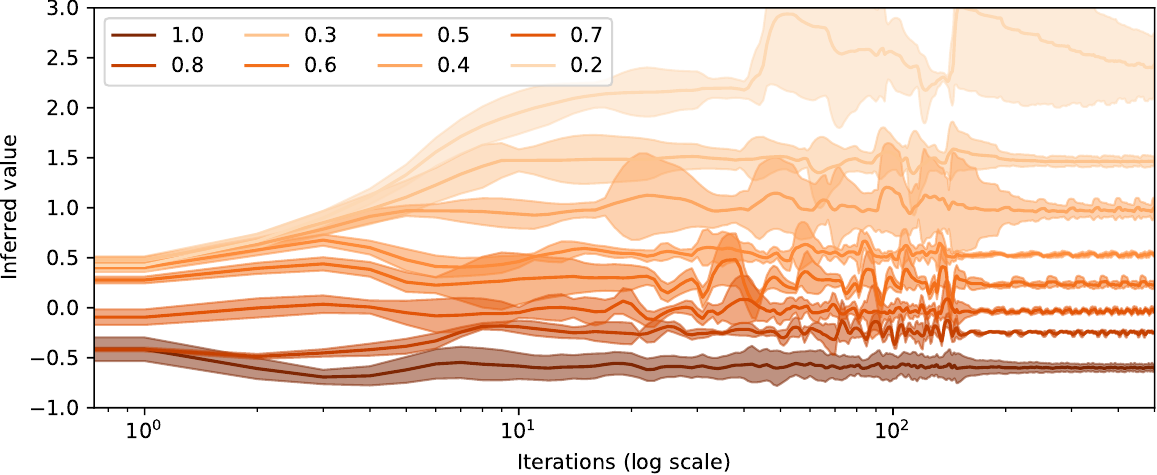}
	\end{minipage}
	\caption{Top left: ground truth and model outputs at time step 80, which is 30 time steps after the start of closed loop prediction (from left to right: ground truth, ConvLSTM, TCN, DISTANA, DISTANA + AT). Bottom left: ground truth and model outputs over time at position $x=2$, $y=6$ in the two-dimensional grid. Top right: inferred static context during testing with values in the range $[-0.6, 2.6]$ after 1, 2, 10, 500 iterations and ground truth with values in $[0.0, 1.0]$. For the ground truth, darker color corresponds to a stronger blocking effect on the wave, which was learned and inferred inversely by the network. Bottom right: average inferred contexts over time during testing (x-axis log scaled).}
	\label{experiments_and_results:fig:context_inference}
\end{figure}

%\begin{table}[!b]
%	\begin{tabularx}{\linewidth}{lCc}  %l@{\extracolsep{\fill}}c@{\extracolsep{\fill}}c}
%		\toprule
%		{Model} & {Parameters} & {MSE}\\
%		\midrule
%		{ConvLSTM} & $2\,952$ & \SI{9.08(43)e-4}{}\\
%		{TCN} & $2\,306$ & \SI{2.94(238)e-2}{}\\
%		{DISTANA} & $1\,236$ & \SI{8.69(87)e-4}{}\\
%		{DISTANA + AT} & $1\,486$ & \SI{3.87(248)e-4}{}\\
%		\bottomrule
%	\end{tabularx}
%	\caption{MSE results at test time confirm the importance of inferring the unknown static context (DISTANA + AT) on the 2D circular wave benchmark.}
%	\label{experiments_and_results:tab:context_inference_mse}
%\end{table}

Furthermore, as shown in \autoref{experiments_and_results:fig:context_inference} (bottom right), DISTANA + AT preserves a linear ordering when inferring context values that were never encountered during training as indicated by the static context values $0.4$ and $0.7$, which are properly mapped to roughly $-0.1$ and $-1.1$, respectively, without violating the propagation speed order with respect to other static context values.
Thus, looking at the estimated static context $\hat{\mathbf{s}}$, it turns out that the latent state inferred by AT correctly reproduced the monotonicity of the here known underlying structure.
The static context map at test time, which is different to the map on which the model was trained on, is inferred correctly (see image sequence of \autoref{experiments_and_results:fig:context_inference}, top right).
When comparing the prediction accuracy of DISTANA and DISTANA + AT in \autoref{experiments_and_results:fig:context_inference} (top and bottom left), the self-inferred static context clearly helps DISTANA + AT to model the two-dimensional wave.

\subsection{WeatherBench}\label{experiments:subsec:weatherbench}

Recently, \cite{rasp2020weatherbench} introduced a benchmark for comparing mid-range (that is three to five days) weather forecast qualities of data driven and physics-based approaches.
While globally regularly aggregated data are provided in three spatial resolutions ($5.625^\circ$, $2.8125^\circ$ and $1.40525^\circ$ resulting in $32\times64$, $64\times128$ and $128\times256$ grid points, respectively), evaluated baselines are reported for the coarsest resolution only, which in consequence we chose too for elaborating and comparing DISTANA.
Baselines are generated by means of persistence (tomorrow's weather is today's weather), climatology, linear regression, and physics-based numerical weather prediction models.
Moreover, convolutional neural networks (CNNs) are either applied iteratively or directly.
Baselines are computed solely on three or five day predictions of the geopotential at an atmospheric pressure level of \SI{500}{hPa} (roughly at \SI{5.5}{km} height, called Z500) and the temperature at \SI{850}{hPa} ($\sim$\SI{1.5}{km} height, referred to as T850).
Beyond Z500 and T850, weatherBench consists of numerous additional dynamic variables (humidity, precipitation, wind direction and speed, solar radiation, etc.), partially reported on multiple vertical layers, and static variables (land-sea mask, soil type, orography, latitude and longitude).

%\def\nlat{N_{\smash{{lat}}}}
%\def\nlong{N_{\smash{{lon}}}}

%For training and testing, a latitude weighted root mean squared error was applied, as proposed in \cite{rasp2020weatherbench}:
%
%\begin{align}
%\small
%\operatorname{RMSE} & = \sqrt{\frac{1}{\nlat\nlong}\sum_{i}^{\nlat}\sum_{j}^{\nlong}L(i)(\hat{y}_{i,j}-y_{i,j})^2}\\
%L(i) & = \displaystyle\cos(\operatorname{lat}(i))\bigg/\left(\displaystyle\frac{1}{\nlat}\sum_{i^\prime}^{\nlat}\cos(\operatorname{lat}(i^\prime))\right),
%\end{align}

%
%where $\nlat$ (latitude) and $\nlong$ (longitude) are the rows and columns of the two-dimensional lattice. 
%In our case ${\nlat=32}$ and ${\nlong64}$, $y_{i,j}$ and $\hat{y}_{i,j}$ are the target and the model's predicted value at the grid position $(i, j)$ and $L(i)$ is the latitude weighting factor, which is largest at the equator (where $\operatorname{lat}(i) = 0$) and decreases towards the poles (where $\operatorname{lat}(i)\rightarrow\pm\frac{\pi}{2}$), such that prediction errors in the equatorial regions -- where the data distortions, coming from transforming the spherical data into a regular rectangular form, are minimal -- contribute stronger to the overall error compared to prediction errors at the poles.

We use weatherBench (a) to explore DISTANA's abilities to approximate real-world phenomena by comparing it to \cite{rasp2020weatherbench}'s iterative CNN approach and (b) to investigate how to apply gradient-based inference techniques in order to infer local static context (e.g. the land-sea mask) that affect Z500 and T850. The experiments we conducted on weatherBench focused on the prediction of the Z500 (geopotential) and T850 (temperature) variables.
DISTANA and DISTANA + AT were trained for $2\,000$ epochs on weather data from 1979, using a learning rate of \SI{e-4}{}, validated on 2016, and tested on 2017.
Each year was partitioned into sequences of 96 hourly steps, yielding 91 sequences per year.
Increasing the set sizes or changing the training, validation, or testing years did not seem to alter the results or model performances.
% which is in accordance with observations from other GNN research, where these models are observed to hardly ever overfit \cite{karlbauer2019distributed}, which is likely rooted in the principle of sharing weights.
DISTANA's preprocessing and LSTM layers were set to 50 neurons and cells, respectively.
Furthermore, the implementation of DISTANA was enhanced to support a varying lateral communication vector size, which then was increased from one to five neurons, to enable neighboring PKs to exchange information of higher complexity, yielding $\sim25\,000$ parameters, slightly varying with the number of input variables.
Moreover, the lateral connection scheme of DISTANA was specified such that information exiting the horizontal boundaries would enter at the other end of the field to match weatherBench's horizontally connected spherical data composition.

Selected static information provided by weatherBench was adapted and extended to facilitate the learning process.
Changes were made to the latitude and longitude variables: latitude was transformed to be zero at the equator and non-linearly rising to one towards the poles, based on $\cos(\text{lat})$.
The longitude variable was split into its sine and cosine component, creating a circular encoding to match the spherical shape of the Earth from which the data origins.
Additionally, one-dimensional north- and south-flags were provided to account for the missing neighbors in the north- and south-most rows in the grid.
As has been done in \cite{weyn2020improving}, we also provide the top of atmosphere total incident solar radiation (tisr).
All variables were normalized to the range of $[-1.0, 1.0]$.
When using AT to infer a latent static context $\tilde{\mathbf{s}}$, the values were clamped to $[-1.0, 1.0]$ to prevent them from drifting or exploding.
If not specified differently, we provide the models with the dynamic variable Z500 or T850 (being subject for prediction), along with nine static inputs: orography, land-sea mask (LSM), soil type, longitude (two-dimensional), latitude, tisr, and the north- and south-end flags.

\subsection{WeatherBench Results}

\begin{figure}
	\centering
	\includegraphics[width=\textwidth]{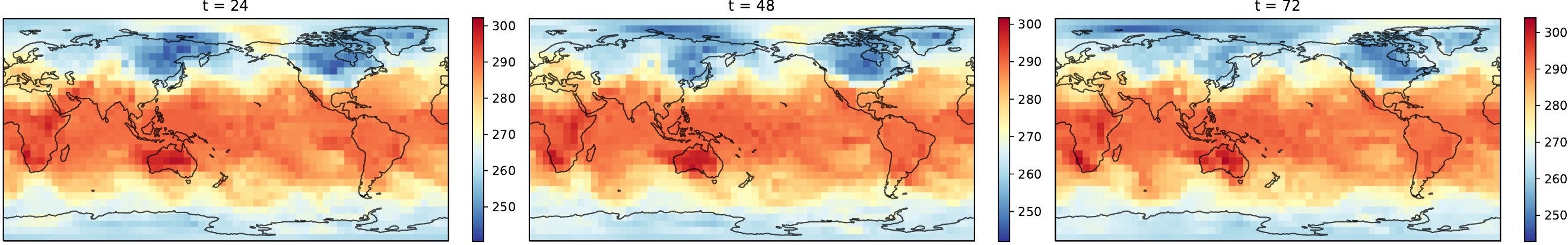}\\
	\vspace{0.08cm}
	\includegraphics[width=\textwidth]{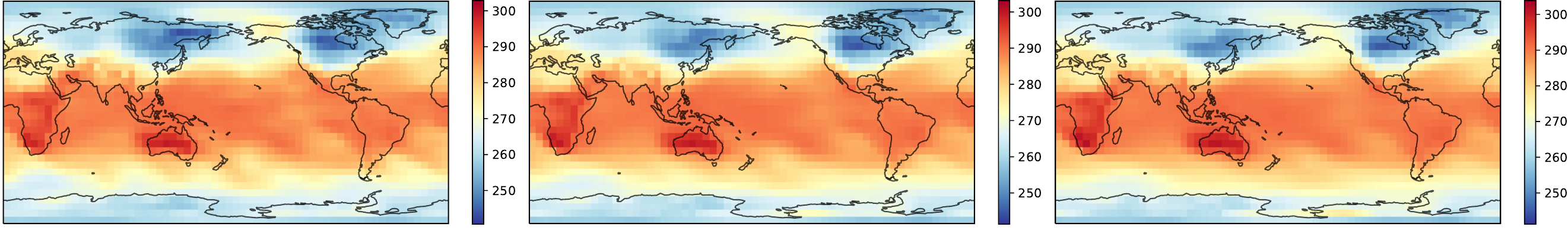}
	\vspace{-0.5cm}
	\caption{Predicted temperature (T850) in degree Kelvin for 24, 48 and 72 hours (corresponding to time steps) into the future (closed-loop). The first row shows the ground truth and the second row the network output.}
	\label{experiments_and_results:fig:t850_prediction}
\end{figure}

%DISTANA was trained on real-world data for the first time, by providing the previously described nine-dimensional static input vector along with the dynamic Z500 variable for prediction.

The evaluation of DISTANA being trained to predict the Z500 variable for a lead time of \SI{72}{h} yielded an RSME of 816, which is better than the current best comparable iterative approach reported on the benchmark ($\text{RMSE}=1114$). 
However, seeing that the best numerical operational weather prediction model produces an RMSE of 154 and other machine learning approaches achieve an RMSE of 268, there is certainly room for improvement. 
Nonetheless, DISTANA offers the best learned generative, iterative processing model on the benchmark without applying techniques that reduce the distortion resulting from transforming the spherical Earth data to a regular two-dimensional grid.

\begin{figure}%[b!]
	\centering
	\includegraphics[width=0.329\textwidth]{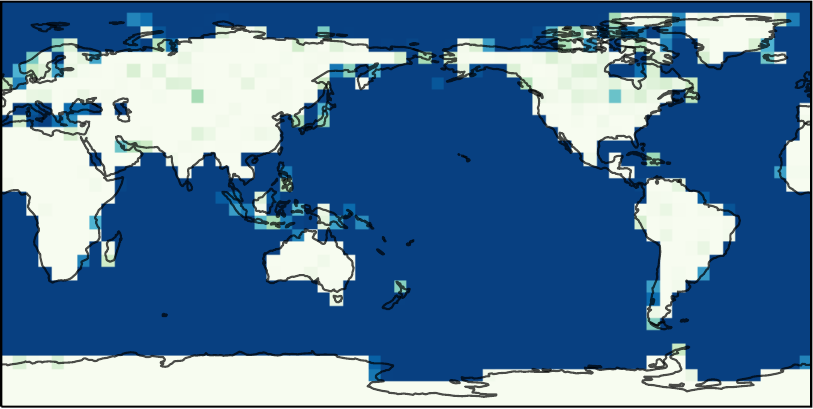}\hfill\includegraphics[width=0.329\textwidth]{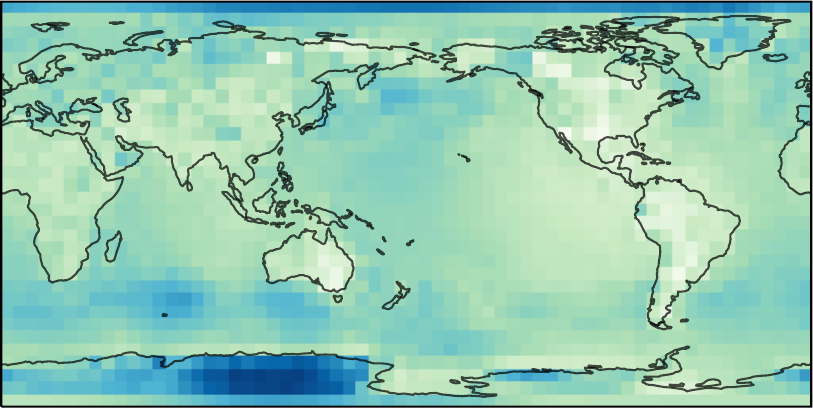}\hfill\includegraphics[width=0.329\textwidth]{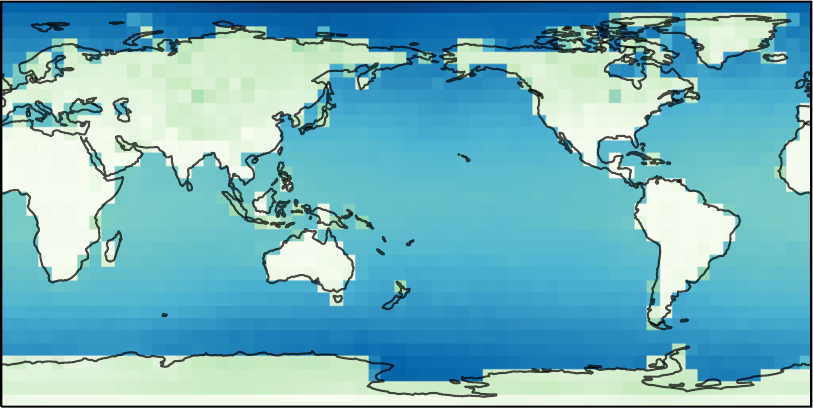}\\
	\vspace{0.08cm}
	\includegraphics[width=0.329\textwidth]{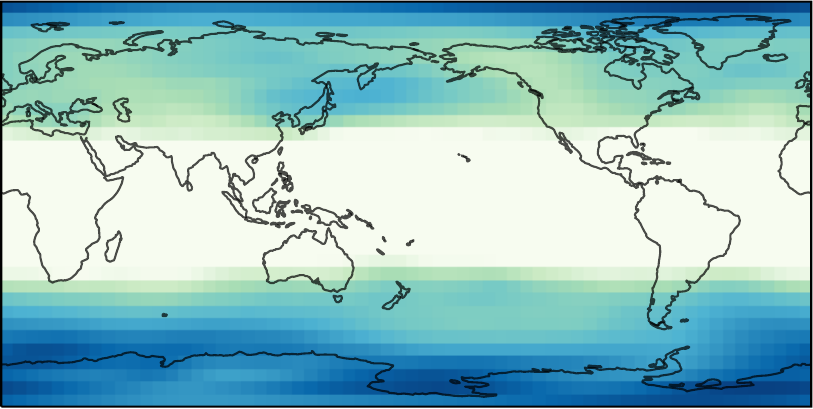}\hfill\includegraphics[width=0.329\textwidth]{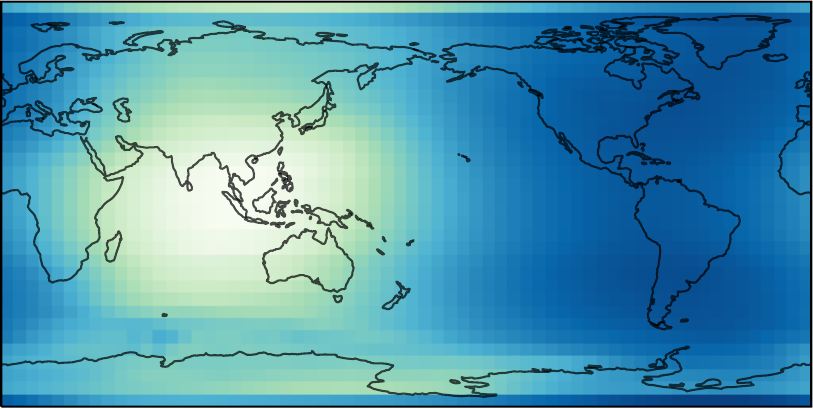}\hfill\includegraphics[width=0.329\textwidth]{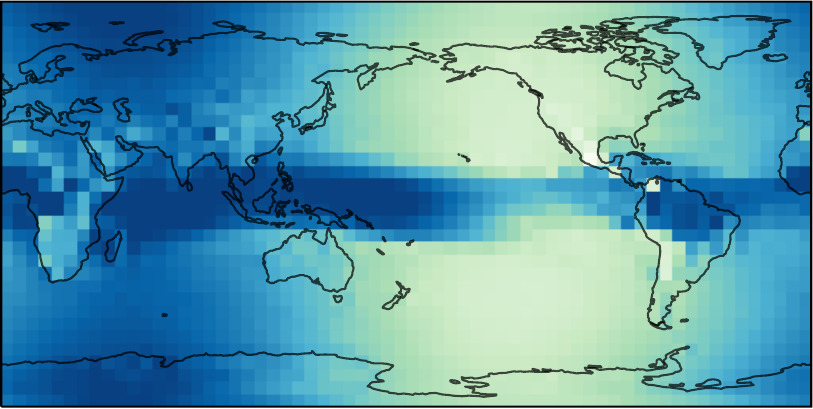}
	\caption{Top left: original land-sea mask (LSM). Top center: global LSM inferred during testing after being trained on two thirds of the globe (the model has never seen America's LSM). Top right: a latent vector which developed during training and encodes LSM information as well as a decent latitude coding. Bottom: three latent variable codes that freely emerged during training of the Z500 (left, center) and T850 (right) variables.}
	\label{experiments_and_results:fig:lsm_inference}
\end{figure}

A second experiment was conducted to (a) investigate whether DISTANA + AT is able to predict the T850 variable, see \autoref{experiments_and_results:fig:t850_prediction}, and (b) simultaneously infer missing land-sea mask (LSM) values only from the observed T850 dynamics.
The model thus received the same static input as in the previous experiment along with the T850 variable during training.
However, only two thirds of the LSM values were provided.
The other third, considered missing values, which covered America and the Atlantic ocean, were to be inferred.
After training, the entire LSM vector $\hat{\mathbf{s}}_\text{LSM}$, initialized with zeros, was retrospectively tuned via AT such that it would best explain the observed dynamics.
As visualized in \autoref{experiments_and_results:fig:lsm_inference} (top center) the missing LSM is inferred reasonably, including the American continent, which the network has never seen during training or inference. These findings suggest that the model learned a generalizable, globally applicable encoding of the LSM's influence on the T850 dynamics.

In a third experiment, we used an additional latent neuron -- a parametric bias neuron -- that is locally tuned during training via AT.
This latent neuron is supposed to be tuned freely by the model to develop any code that helps the model to predict the observed dynamics.
%Numerous conditions were evaluated, by either training to predict the Z500 or the T850 variable and providing the network with or without land-coding variables (e.g. orography, LSM, soil type).
We were particularly interested in evaluating whether DISTANA would develop latent states $\tilde{\mathbf{s}}$ that distinctively encode prediction-relevant, hidden causal factors that correspond to observable values.
For example, we wanted to see whether DISTANA would develop a latent code that resembles any land-coding quantity. 
Thus, in this experiment, we try to answer the question what latent states are inferred depending on the predicted variable and how the presence of land-relevant input does affect the generation of this latent code.

Our results indicate that the nature of the developed latent states depends considerably on both the variable that is subject for prediction (Z500 or T850) and the additional static data provided.
\autoref{experiments_and_results:fig:lsm_inference} (top right) shows a clear tendency to encode land-sea information, augmenting it with a latitude code, when all previously mentioned static inputs (including LSM) were provided.
When training a model to predict Z500, the emerging latent variables rather seem to encode latitude, albedo, monsoon \cite{boers2019complex}, or humidity-distribution patterns (\autoref{experiments_and_results:fig:lsm_inference} bottom left and center).
Excitingly, nuances of LSM and orography become visible when training to predict T850 without receiving any land-coding inputs (see \autoref{experiments_and_results:fig:lsm_inference} bottom right).
%In other training runs, the inferred latent variable seemed to encode influences due to albedo, monsoon (compare \cite{boers2019complex}), and global humidity distribution patterns, as visualized in the appendix of the supplementary material -- further studies are necessary to verify these observations in detail.
%Qualitative results on the spatiotemporal prediction of the Z500 and T850 variables along with video material is provided in the supplementary material.
Nevertheless, further studies are necessary to verify to which extent the inferred variables correlate with observations in detail.
%It is also important to note that, when local latent variables $\tilde{\mathbf{s}}$ are included to develop freely during training and testing, the prediction error of the network decreases slightly ($\approx 5-10\%$), but reliably.

%%%%%%%%%%%%%%%%%%%%%%%%%%%%%%%%%%%%%%%%
%%%%%%%%%%%%%% DISCUSSION %%%%%%%%%%%%%%
%%%%%%%%%%%%%%%%%%%%%%%%%%%%%%%%%%%%%%%%

\section{Final Discussion}
The presented results indicate that the combination of the DIstributed, SpatioTemporal graph Artificial Neural network Architecture, DISTANA, with the retrospective inference mechanism called active tuning (AT), bears large potential at predicting spatiotemporal real-world phenomena (e.g. weather). 
It outperforms competing deep learning algorithms by generating more accurate closed-loop predictions into the future.
In addition, it can infer hidden causes by mere observation of a dynamic process.
In particular, AT in DISTANA is well-suited for inferring (i) contrastive hidden causes during learning and (ii) hidden static activities while minimizing loss online. 
%AT optimizes neural activities by means of backpropagation through time, aiming at minimizing cumulative predictive loss.
%As a result, network activities in form of latent codes get tuned in to the data stream and hidden, dynamics-influencing factors can be identified.
%Note that the network is never explicitly informed about the values of any of these static factors, which prevents the application of conventional supervised learning.
While we believe that these hidden factors tend to identify causal influences -- because they form for improving the accuracy of the predicted dynamics -- future research will need to investigate the robustness of this tendency.

During learning, cumulative error signals in latent parametric bias neurons at the individual prediction kernels tend to develop encodings of hidden, dynamic-influencing factors.
To a certain extent, these neuronal encodings resemble physical properties, such as albedo or the land-sea mask, depending on the type of dynamics that is to be predicted (e.g. temperature or geopotential).
%During online prediction, when the network ideally is able to generate all possible dynamics in the analyzed data stream, the retrospective inference of recurrent, dynamic neural activities tunes the network dynamics in to the observed data streams. 
That is, the projection of the gradient onto static neural activities identifies local parametric bias activities that best characterize local, hidden causal factors. 
%When projecting the gradient signal onto locally and temporally static neural activities, we essentially assume that hidden local factors exist that are constant and influence the unfolding spatiotemporal dynamics in a systematic manner. 
%In the future, other assumptions about these hidden factors may be applied, such as a cyclic impulse, a basic wave pattern, or smoothly changing activities. 

%Similar abilities were shown, for example, when learning to control different 2D vehicle dynamics \cite{Butz:2019a}, or when analyzing distinct sensorimotor interaction streams \cite{Sugita:2011a}.
%When this technique is applied in DISTANA, where the weights are shared by all PKs, the bias neurons develop universally applicable prediction dynamics modifiers.
%Under the assumption that these factors are locally constant, these modifiers tend to identify hidden values of local dynamics-influencing causal factors, or blends thereof. 
% Given these hidden states have learned to encode particular causal factors, the factor values can be inferred even if locally unknown. 

Overall, the results suggest that our approach of assuming and inferring hidden causes with constrained properties -- such as being locally distinct, constant, but universally present -- offers strong potential in fostering the development of process-explaining structures. 
%Seeing that the introduced methods are generally complementary to standard training techniques and architectural network design, we are convinced that the application of suitable modifications of the introduced techniques to other domains still holds large potential. 
%Future research will show if this conjecture is correct.

\bibliographystyle{splncs04}
\bibliography{literature}

\end{document}